\documentclass{article} 
\usepackage{iclr2021_conference,times}

\usepackage{amsmath,amsfonts,bm}

\def\eqref#1{equation~\ref{#1}}

\def\1{\bm{1}}

\DeclareMathAlphabet{\mathsfit}{\encodingdefault}{\sfdefault}{m}{sl}
\SetMathAlphabet{\mathsfit}{bold}{\encodingdefault}{\sfdefault}{bx}{n}

\DeclareMathOperator*{\argmax}{arg\,max}

\usepackage{hyperref}
\usepackage{url}

\usepackage{graphicx}
\usepackage{amsmath}
\usepackage{amssymb}
\usepackage{gensymb}

\usepackage{amssymb}
\usepackage{xcolor}

\usepackage{subcaption}

\usepackage{kotex}
\usepackage{color}
\usepackage{array}
\usepackage{hyperref}

\usepackage{bm}
\usepackage{nth}
\usepackage{setspace}

\newcolumntype{C}[1]{>{\centering\let\newline\\\arraybackslash\hspace{0pt}}m{#1}}
\newcolumntype{L}[1]{>{\let\newline\\\arraybackslash\hspace{0pt}}m{#1}}

\usepackage{rotating} 

\usepackage{amsmath}

\title{Test-Time Mixup Augmentation for Data and Class-Specific Uncertainty Estimation in Deep Learning Image Classification}

\author{Hansang Lee \& Haeil Lee\\
School of Electrical Engineering\\
Korea Advanced Institute of Science and Technology\\
Daehark 291, Yuseonggu, Daejeon 34141, Republic of Korea\\
\texttt{\{hansanglee,haeil.lee\}@kaist.ac.kr}\\
\AND 
Helen Hong \thanks{Corresponding author}\\
Department of Software Convergence\\
College of Interdisciplinary Studies for Emerging Industries\\
Seoul Women’s University\\
Hwarangro 621, Nowongu, Seoul 01797, Republic of Korea\\
\texttt{hlhong@swu.ac.kr}\\
\AND
Junmo Kim\\
School of Electrical Engineering\\
Korea Advanced Institute of Science and Technology\\
Daehark 291, Yuseonggu, Daejeon 34141, Republic of Korea\\
\texttt{junmo.kim@kaist.ac.kr}\\
}

\iclrfinalcopy 
\begin{document}

\maketitle

\begin{abstract}
Uncertainty estimation of trained deep learning networks is valuable for optimizing learning efficiency and evaluating the reliability of network predictions. 
In this paper, we propose a method for estimating uncertainty in deep learning image classification using test-time mixup augmentation (TTMA). 
To improve the ability to distinguish correct and incorrect predictions in existing aleatoric uncertainty, we introduce TTMA data uncertainty (TTMA-DU) by applying mixup augmentation to test data and measuring the entropy of the predicted label histogram.
In addition to TTMA-DU, we propose TTMA class-specific uncertainty (TTMA-CSU), which captures aleatoric uncertainty specific to individual classes and provides insight into class confusion and class similarity within the trained network. 
We validate our proposed methods on the ISIC-18 skin lesion diagnosis dataset and the CIFAR-100 real-world image classification dataset. 
Our experiments show that (1) TTMA-DU more effectively differentiates correct and incorrect predictions compared to existing uncertainty measures due to mixup perturbation, and (2) TTMA-CSU provides information on class confusion and class similarity for both datasets.
\end{abstract}



\section{Introduction}
\label{sec:intro}

Uncertainty estimation is a fundamental task in machine learning, offering insights into the reliability and trustworthiness of a model's predictions~\citep{Gawlikowski2021}.
In real-world applications, where decisions based on these predictions can have significant consequences, understanding uncertainty is crucial. 
Aleatoric uncertainty, which captures the inherent noise and variability in the observed data, is vital in various scenarios in medical diagnosis~\citep{Martin2019,Ayhan2020,Cicalese2021,Carneiro2020,Singh2020,Herzog2020,Shamsi2021,Wang2021,Czolbe2021,Matsunaga2017,Graham2019}.
Effective techniques for estimating aleatoric uncertainty can improve decision-making by indicating when predictions may be susceptible to high variability~\citep{Gal2017,Zhao2021,Hong2020,Rizve2021,Nielsen2019}.

Several methods have been proposed to quantify uncertainties in deep learning models in recent years.
Bayesian neural network (BNN) is a popular approach for modeling both aleatoric and epistemic uncertainties, but it can be computationally expensive and complex to implement~\citep{Kendall2017}. 
Monte Carlo dropout (MCDO) simulates a Bayesian approximation, but is more applicable to epistemic uncertainty~\citep{Gal2016}.
More recently, test-time data augmentation (TTA) has gained attention as an approach to estimating aleatoric uncertainties~\citep{Wang2019}. 
This method involves producing multiple augmented versions of a single test data, predicting with each version, and then quantifying the range of predictions to measure uncertainty.
The TTA-based uncertainty has shown promising performance with traditional augmentation techniques, such as flipping, rotation, and scaling, in various image classification tasks~\citep{Moshkov2020TTACellSeg,Wang2019,Nalepa2020TTARemoteSensing}.

Mixup is a data augmentation technique that generates synthetic data points by blending two data instances and their corresponding labels~\citep{Zhang2018}.
The mixup has proven to be an effective tool for training, enhancing both model generalization and robustness. 
Due to its straightforward implementation and effectiveness, it stands as a valuable tool for deep learning across wide range of applications, such as image recognition~\citep{Yun2019,resizemix,Kim2020,saliencymix,supermix}, semantic segmentation~\citep{msdaseg1,msdaseg2,msdaseg3,msdaseg4,msdaseg5}, natural language processing~\citep{msdanlp1,msdanlp3,msdanlp3,msdanlp4,msdanlp5,msdanlp6}, video processing~\citep{msdavideo1,msdavideo2,msdavideo3,msdavideo4}, and medical image analysis~\citep{msdamed1,msdamed2,msdamed3,msdamed4,msdamed5,Verma2018,msdamed7}. 
However, the potential of mixup in the realm of uncertainty estimation, especially during test time, remains largely unexplored.

In this paper, we propose a test-time mixup augmentation (TTMA) method for robust and effective aleatoric uncertainty estimation in image classification. 
Our method includes two distinct uncertainty measures: TTMA data uncertainty (TTMA-DU) and TTMA class-specific uncertainty (TTMA-CSU). 
As other aleatoric uncertainty measures, TTMA-DU is computed by applying mixup to the target test data with uniformly sampled data, making predictions with each version, and then quantifying the entropy of predictions.
By evaluating the model's prediction variability over a range of mixup-augmented test data, TTMA-DU can offer better evaluation of the trained network's reliability compared to the conventional aleatoric uncertainty methods.
On the other hand, TTMA-CSU is computed by applying mixup to the target test data with a specific class data, making predictions with each version, and then quantifying the entropy of predictions.
Unlike TTMA-DU and existing aleatoric uncertainty measures, which are computed independently of class, TTMA-CSU is determined dependently on both the target test data and the specific class.
As a novel type of aleatoric uncertainty measure associated with a specific class, TTMA-CSU can provide insight into the class confusion and class similarity of the trained network.
To evaluate the effectiveness of the proposed methods, we conduct experiments on two publicly available image classification datasets with different characteristics: ISIC-18 and CIFAR-100.
Experiments on both datasets demonstrate that (1) the proposed TTMA-DU yields improved aleatoric uncertainty measures compared to the conventional TTA and MCDO methods, and (2) the proposed TTMA-CSU provides insights into class confusion and class similarity in the latent space of the trained network.

The main contributions of this work can be summarized as follows.
\begin{itemize}
    \item We propose \textit{TTMA data uncertainty (TTMA-DU)}, a novel aleatoric uncertainty measure that outperforms existing TTA uncertainty in evaluating the reliability of network predictions.
    \item We introduce \textit{TTMA class-specific uncertainty (TTMA-CSU)}, the first aleatoric uncertainty measure associated with a specific class, which provides insights into class confusion and similarity within the trained network.
    \item We validate the effectiveness of these uncertainty measures through experiments conducted on two publicly available image classification datasets with diverse characteristics.
\end{itemize}

\section{Methods}
\label{sec:method}

\begin{figure*}[t!]
\centering
\includegraphics[width=\textwidth]{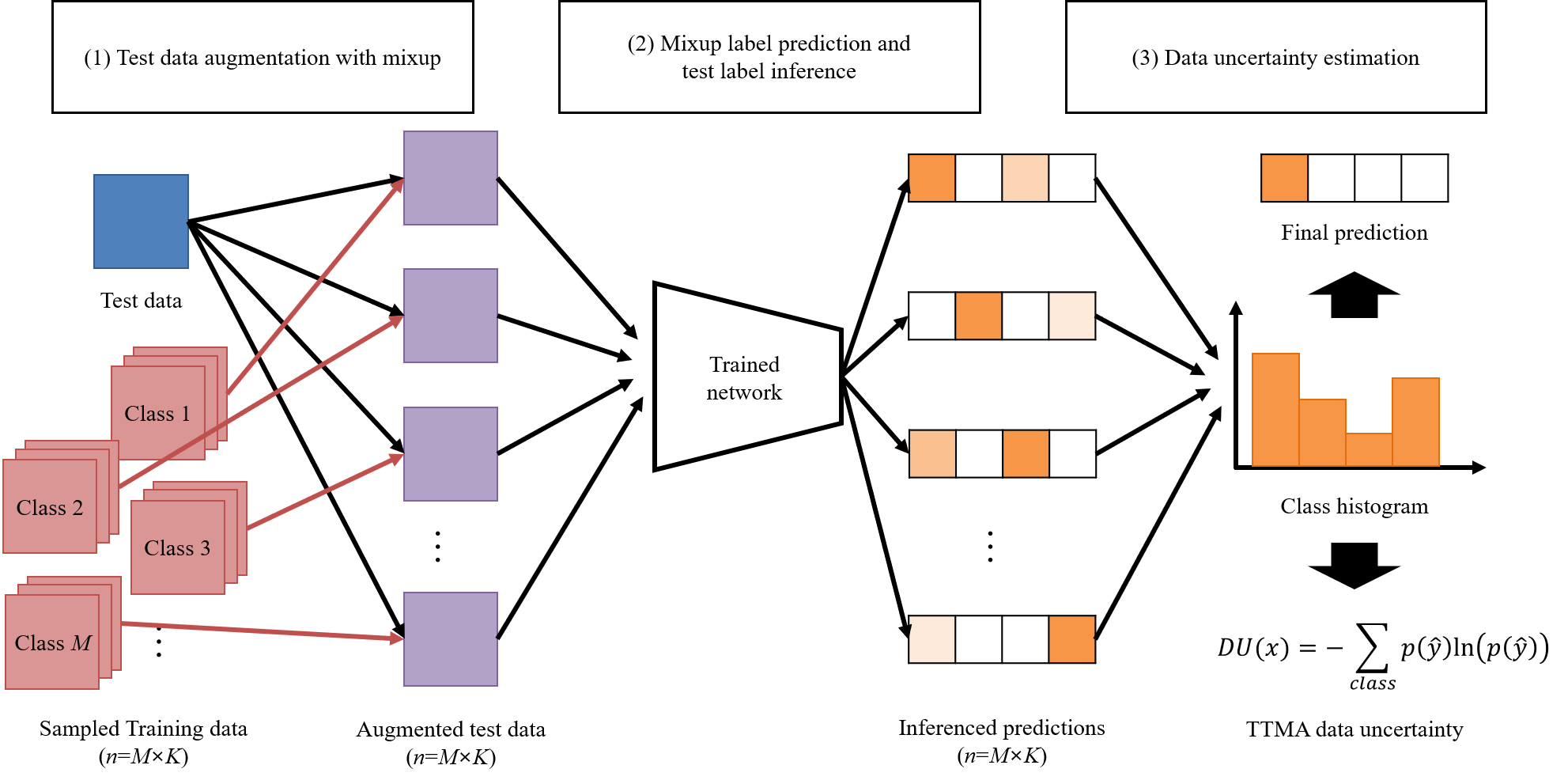} 
\caption{A process of the proposed TTMA data uncertainty (TTMA-DU) estimation method.}
\label{fig:method_du}
\end{figure*}
\subsection{TTMA data uncertainty}
\label{sec:dataunct}

The proposed \textit{TTMA data uncertainty (TTMA-DU)} aims to enhance TTA uncertainty by replacing the affine augmentation with mixup augmentation for test data.
The proposed method consists of three steps: (1) test data augmentation with mixup, (2) mixup label prediction and test label inference, and (3) TTMA-DU estimation.
The process of estimating TTMA-DU is illustrated in Fig.~\ref{fig:method_du}.

\subsubsection{Test data augmentation with mixup}

First, we apply mixup augmentation to the test data to obtain perturbation-robust results and to estimate the uncertainty.
Let us have a training image-label pair $(x_{train},y_{train})\in S_{train}$, and a test image-label pair $(x_{test},y_{test})\in S_{test}$, where $x\in \mathbb{R}^{W \times H}$ is an input image with width of $W$ and height of $H$, and $y \in \mathbb{R}^{M}$ is a soft label vector with a size of $M$ classes.
For a given test data $(x_{test},y_{test})$, we form a mixup test data $(x_{mixup},y_{mixup})$ by combining the test data with randomly sampled training data $(x_{train},y_{train})$ as follows.

\begin{equation}
    x_{mixup}(m,k) = \lambda x_{test} + (1-\lambda) x_{train}(m,k)
    \label{eq:1}
\end{equation}

\begin{equation}
    y_{mixup}(m,k) = \lambda y_{test} + (1-\lambda) y_{train}(m,k)
    \label{eq:2}
\end{equation}

\noindent
where $(x_{train}(m,k),y_{train}(m,k))$ is a training image-label pair of $k$-th data randomly sampled from class $m \in \{1,...,M\}$, and $\lambda$ is a mixup coefficient determined by the beta distribution variable $\lambda \sim \mathcal{B}(\alpha,\alpha)$. This process generates a total of $M \times K$ mixup-augmented data for one test data, where $K$ is a number of sampled training data.


\subsubsection{Mixup label prediction and test label inference}

In this step, we utilize the mixup-trained network $f:\mathbb{R}^{W\times H}\rightarrow \mathbb{R}^{M}$, which has been trained on the training set $S_{train}$, to make predictions for the soft labels of the mixup-augmented test data $\hat{y}_{mixup}(m,k)=f(x_{mixup}(m,k))$.
Subsequently, the label of the test data $\hat{y}_{test}(m,k)$ is inferred from this prediction result.
From Eq. \ref{eq:2}, we have

\begin{equation}
    y_{test} = \left( y_{mixup}(m,k) - (1-\lambda)y_{train}(m,k) \right) /\lambda
    \label{eq:3}
\end{equation}

By replacing the true labels $y_{test}, y_{mixup}(m,k)$ with the inferred label of the test data $\hat{y}_{test}(m,k)$ and the predicted label of the mixup-augmented data $\hat{y}_{mixup}(m,k) = f(x_{mixup}(m,k))$, respectively, we can rewrite Eq.~\ref{eq:3} as

\begin{equation}
    \hat{y}_{test}(m,k) = \left( f(x_{mixup}(m,k)) - (1-\lambda)y_{train}(m,k) \right) /\lambda
    \label{eq:4}
\end{equation}

\noindent
From Eq.~\ref{eq:4}, we can obtain a total of $M \times K$ inferred labels for one test data.

\subsubsection{TTMA data uncertainty estimation}

From a total of $M \times K$ inferred soft labels for the test data $\hat{y}_{test}(m,k)$, we can have a histogram of hard labels $\hat{\mathcal{Y}}_{test} = \{ \argmax_{l} \hat{y}_{test}(m,k), m=1,...,M, k=1,...,K \}$ where $l=1,...,M$ is a class index for hard label.
This histogram of inferred hard labels $\hat{\mathcal{Y}}_{test}$ is then used to determine the final test label $\hat{y}_{test}$ through a majority voting as follows:

\begin{equation}
    \hat{y}_{test} = \argmax_{l}P_{l}\left(\hat{\mathcal{Y}}_{test}\right)
    \label{eq:5}
\end{equation}

\noindent
where $P_{l}(\hat{\mathcal{Y}}_{test})$ is the probability that the histogram of $\hat{\mathcal{Y}}_{test}$ belongs to class $l=1,...,M$.

TTMA-DU is then computed as the entropy of the histogram of inferred labels $\hat{\mathcal{Y}}_{test}$ by

\begin{equation}
    DU(x_{test}) = -\sum_{l=1}^{M}{P_{l}\left(\hat{\mathcal{Y}}_{test}\right) ln\left(P_{l}\left(\hat{\mathcal{Y}}_{test}\right)\right)}.
    \label{eq:6}
\end{equation}

The proposed TTMA-DU illustrates the instability of test data predictions resulting from the mixture of various classes.
It introduces more intense perturbations compared to traditional affine-based transformations, enabling an evaluation of the trained network's robustness to data-based perturbations under more challenging conditions than those offered by existing TTA methods~\citep{Wang2019}.

\subsection{TTMA class-specific uncertainty}
\label{sec:cdunct}

\begin{figure*}[t!]
\centering
\includegraphics[width=\textwidth]{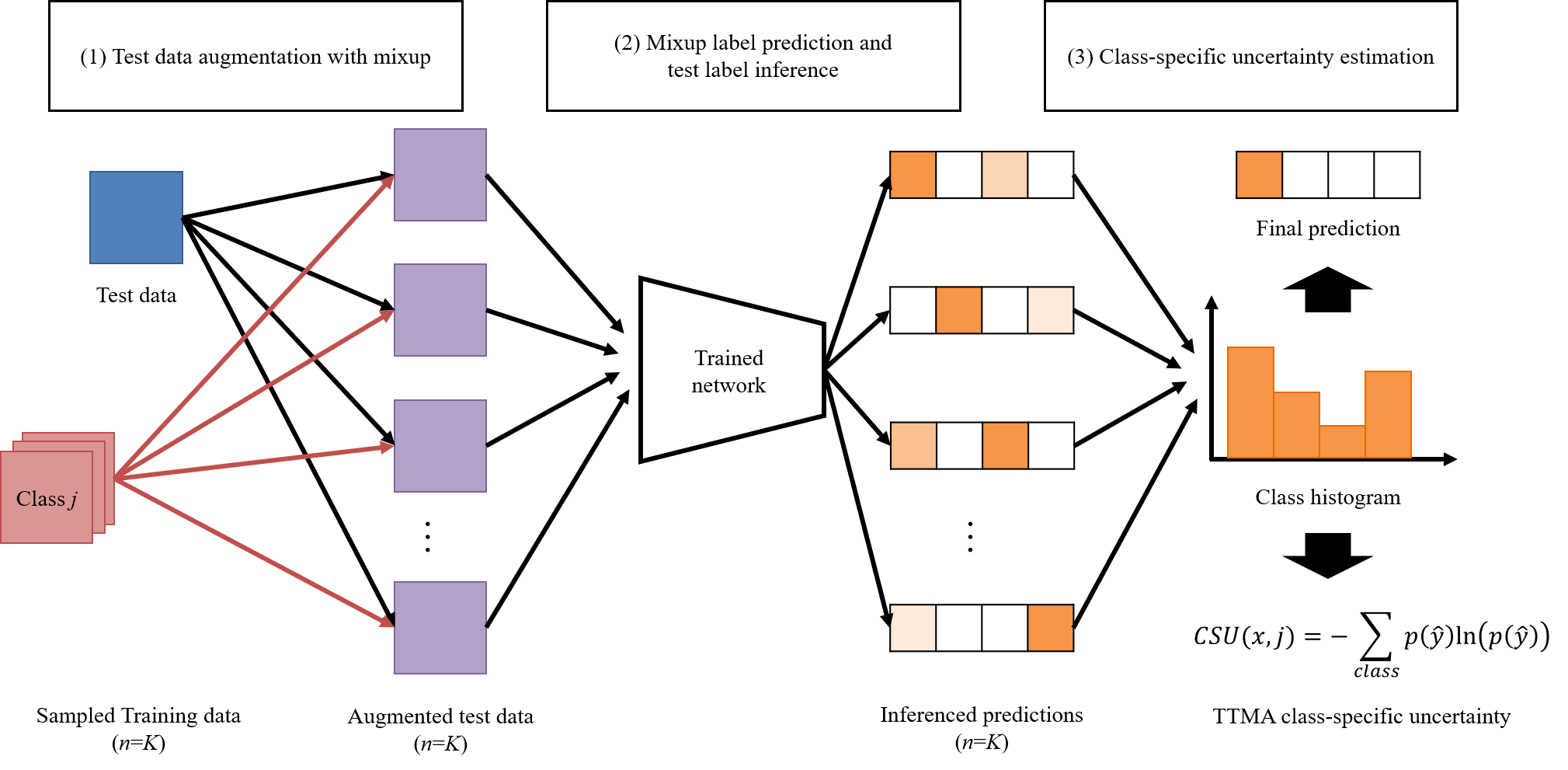} 
\caption{A process of TTMA class-specific (TTMA-CSU) uncertainty estimation method.}
\label{fig:method_cdu}
\end{figure*}

In the computation of TTMA-DU, mixup is performed with the equal number of sampled data from all classes to apply unbiased perturbation to the test data.
However, this raises questions about how the uncertainty measures appear and which information they can provide when they are based on test data perturbed exclusively by mixup with the training data from a specific class.
We refer to this as \textit{TTMA class-specific uncertainty (TTMA-CSU.)}
To address this, the proposed method for estimating TTMA-CSU involves three steps: (1) Test data augmentation with mixup, (2) mixup label prediction and test label inference, and (3) TTMA-CSU estimation. 
The process of estimating TTMA-CSU is illustrated in Fig.~\ref{fig:method_cdu}.

\subsubsection{Test data augmentation with mixup}

First, we apply mixup augmentation to the test data with the training data sampled from a specific class $j \in \{1,...,M\}$ to obtain perturbation-robust results and to estimate the uncertainty.
Let us have a training image-label pair $(x_{train},y_{train})\in S_{train}$, and a test image-label pair $(x_{test},y_{test})\in S_{test}$, where $x\in \mathbb{R}^{W \times H}$ is an input image with width of $W$ and height of $H$, and $y \in \mathbb{R}^{M}$ is a soft label vector with a size of $M$ classes.
For a given test data $(x_{test},y_{test})$, we form a mixup test data $(x_{mixup},y_{mixup})$ by combining the test data with randomly sampled training data $(x_{train},y_{train})$ as follows.

\begin{equation}
    x_{mixup}(j,k) = \lambda x_{test} + (1-\lambda) x_{train}(j,k)
    \label{eq:7}
\end{equation}

\begin{equation}
    y_{mixup}(j,k) = \lambda y_{test} + (1-\lambda) y_{train}(j,k)
    \label{eq:8}
\end{equation}

\noindent
where $(x_{train}(j,k),y_{train}(j,k)$ is a training image-label pair of $k$-th data randomly sampled from training set of class $j$, and $\lambda$ is a mixup coefficient determined by the beta distribution variable $\lambda \sim \mathcal{B}(\alpha,\alpha)$.
This process generates a total of $K$ mixup-augmented data for one test data, where $K$ is a number of sampled training data.

\subsubsection{Mixup label prediction and test label inference}

In this step, we utilize the mixup-trained network $f:\mathbb{R}^{W\times H}\rightarrow \mathbb{R}^{M}$, which has been trained on the training set $S_{train}$, to make predictions for the soft labels of the mixup-augmented test data $\hat{y}_{mixup}(j,k)=f(x_{mixup}(j,k))$.
Subsequently, the label of the test data $\hat{y}_{test}(j,k)$ is inferred from this prediction result.
From Eq. \ref{eq:8}, we have

\begin{equation}
    y_{test} = \left( y_{mixup}(j,k) - (1-\lambda)y_{train}(j,k) \right) /\lambda
    \label{eq:9}
\end{equation}

By replacing the true labels $y_{test}, y_{mixup}(j,k)$ with the inferred label of the test data $\hat{y}_{test}(j,k)$ and the predicted label of the mixup-augmented data $\hat{y}_{mixup}(j,k) = f(x_{mixup}(j,k))$, respectively, we can rewrite Eq.~\ref{eq:9} as

\begin{equation}
    \hat{y}_{test}(j,k) = \left( f(x_{mixup}(j,k)) - (1-\lambda)y_{train}(j,k) \right) /\lambda
    \label{eq:10}
\end{equation}

\noindent
From Eq.~\ref{eq:10}, we obtain a total of $K$ inferred labels for one test data.

\subsubsection{TTMA class-specific uncertainty estimation}

From a total of $K$ inferred soft labels for the test data $\hat{y}_{test}(j,k)$, we can have a histogram of hard labels $\hat{\mathcal{Y}}_{test,j} = \{ \argmax_{l} \hat{y}_{test}(j,k), k=1,...,K \}$ where $l=1,...,M$ is a class index for hard label.
From the histogram of inferred hard labels $\hat{\mathcal{Y}}_{test,j}$, the final test label $\hat{y}_{test,j}$ can be obtained using majority voting by

\begin{equation}
    \hat{y}_{test,j} = \argmax_{l}P_{l}\left(\hat{\mathcal{Y}}_{test,j}\right)
    \label{eq:11}
\end{equation}

\noindent
where $P_{l}(\hat{\mathcal{Y}}_{test,j})$ is a probability that the histogram of $\hat{\mathcal{Y}}_{test,j}$ has a class $l=1,...,M$.

TTMA-CSU is then computed as the entropy of the histogram of inferred labels $\hat{\mathcal{Y}}_{test,j}$ by

\begin{equation}
    CSU(x_{test},j) = -\sum_{l=1}^{M}
    P_{l}\left(\hat{\mathcal{Y}}_{test,j} \right) ln\left(P_{l}\left(\hat{\mathcal{Y}}_{test,j} \right)\right).
    \label{eq:12}
\end{equation}

From the perspective of existing aleatoric uncertainty, TTMA-CSU represents the instability in test data predictions arising from perturbations introduced by mixup with a specific class.
To delve into the specific interpretation and effectiveness of TTMA-CSU, we formulate a hypothesis and verify it through experiments.


\begin{figure}[t!]
\centering
\includegraphics[width=0.4\textwidth]{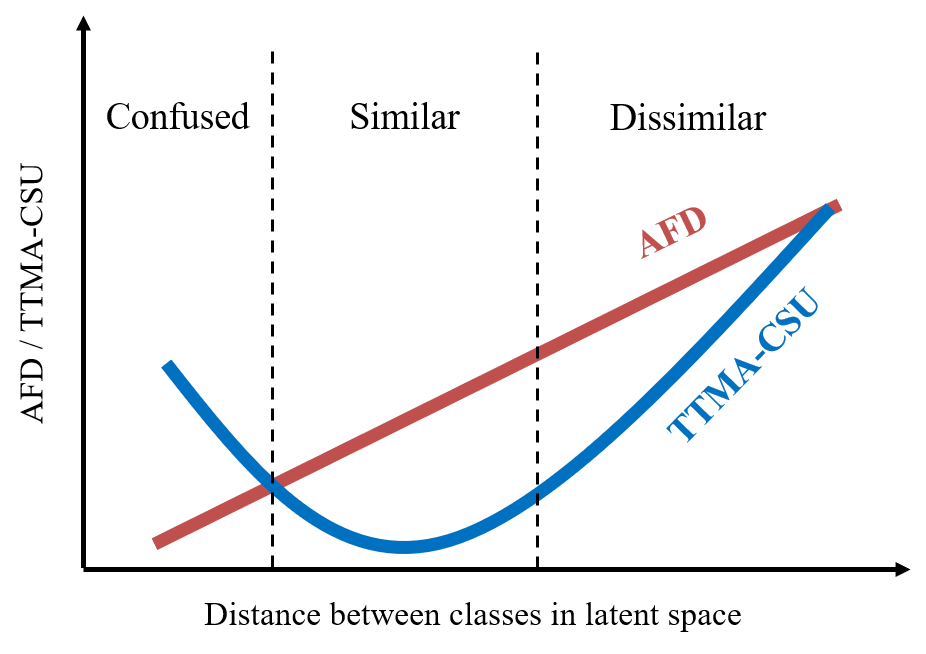} 
\caption{Our hypothesis on TTMA class-specific uncertainty (TTMA-CSU) and average feature distance (AFD) according to class dissimilarity of two classes in class confusion and class similarity scenarios.}
\label{fig:csu_vs_afd}
\end{figure}

\subsubsection{Interpretation of TTMA class-specific uncertainty}

To understand the insights TTMA-CSU offers about the relationship between data and class, we first categorize the relationships between classes based on the distance between the classes in the latent space of a trained network.
First, two classes are \textit{dissimilar} when the distance between them in latent space is large and the network clearly distinguishes them.
Second, two classes are \textit{similar} is the distance is close but the boundary is well learned and the network can sufficiently separate them.
Third, two classes are \textit{confused} each other when the distance is not only close but also the classes almost overlap in the latent space, and the network may struggle to differentiate them.
Identifying these class relationships, particularly those involving similarity and confusion, is crucial when assessing a trained network's performance and potential for improvement.

One conventional measure to quantify the class relationships in the latent space is average feature distance (AFD).
The AFD between the test data $x_{test}$ and the specific class $j$ can be calculated as

\begin{equation}
    AFD(x_{test},j) = \frac{1}{K} \sum_{k=1}^{K} d \left( v_{test}, v_{train}(j,k) \right),
    \label{eq:13}
\end{equation}

\noindent
where $K$ is a number of sampled training data in class $j$, $v_{test}, v_{train}(j,k)$ are feature vectors obtained from the input data $x_{test}, x_{train}(j,k)$ in the network, respectively, and $d(u,v)=1-(u \cdot v)/(\|u\|\|v\|)$ is a cosine distance between $u$ and $v$.

Fig.~\ref{fig:csu_vs_afd} illustrates the change in AFD values for three class relationships.
Given that AFD is monotonically proportional to class distance, the class relationships can be differentiated through suitable thresholds.
However, in both class confusion and class similarity, AFD has low values and the threshold between them is ambiguous, so there is a limitation in effectively distinguishing the two circumstances using only AFD.

In contrast, TTMA-CSU exhibits distinct characteristics compared to AFD regarding the class relationships, as shown in Fig.~\ref{fig:csu_vs_afd}.
For class similarity and dissimilarity cases, both TTMA-CSU and AFD have low and high values, respectively.
However, when two classes are confused, TTMA-CSU increases while AFD remains low.
This is due to the ambiguity about the class involved in the mixup, which results in increasing the instability of predictions.
This contrasting behavior highlights the unique strengths of TTMA-CSU in identifying class confusion and similarity along with AFD.

\section{Experiments}
\label{sec:exp}

\begin{table}[!b]
\caption{Statistics on the amount of data by disease class for the training and validation sets in the ISIC-18 dataset. (AKIEC: Actinic keratosis, BCC: Basal cell carcinoma, BKL: Benign keratosis, DF: Dermatofibroma, MEL: Melanoma, NV: Melanocytic nevus, VASC: Vascular lesion}
\centering
\resizebox{0.6\columnwidth}{!}{
\begin{tabular}{C{3cm} | C{2.5cm} C{2.5cm}}
\hline
\textbf{Classes} & \textbf{Training} & \textbf{Validation} \\
\hline \hline
AKIEC & 327 & 8 \\
BCC & 514 & 15 \\
BKL & 1099 & 22 \\
DF & 115 & 1 \\
MEL & 1113 & 21 \\
NV & 6705 & 123 \\
VASC & 142 & 3 \\
\hline
Total & 10015 & 193 \\
\hline
\end{tabular}
}
\label{table:isic18}
\end{table}

\begin{figure}[!t]
    \centering
        \subfloat[]{\includegraphics[width=0.06\columnwidth]{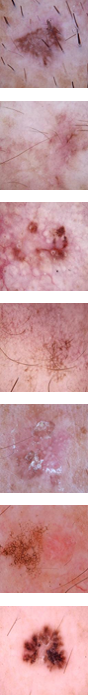}}
        \subfloat[]{\includegraphics[width=0.06\columnwidth]{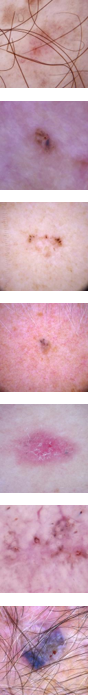}}
        \subfloat[]{\includegraphics[width=0.06\columnwidth]{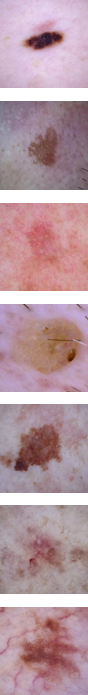}}
        \subfloat[]{\includegraphics[width=0.06\columnwidth]{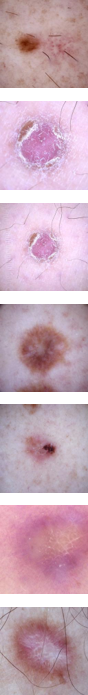}}
        \subfloat[]{\includegraphics[width=0.06\columnwidth]{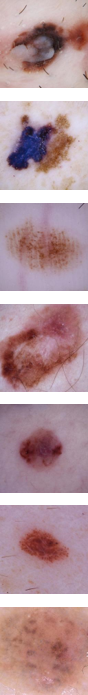}}
        \subfloat[]{\includegraphics[width=0.06\columnwidth]{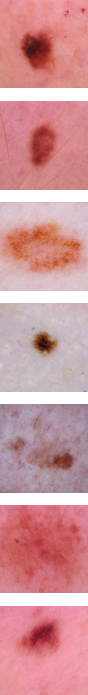}}
        \subfloat[]{\includegraphics[width=0.06\columnwidth]{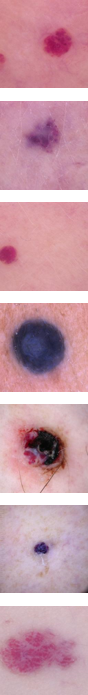}}
    \caption{Examples of ISIC-18 skin lesion images for different disease clases: (a) AKIEC, (b) BCC, (c) BKL, (d) DF, (e) MEL, (f) NV, (g) VASC.}
    \label{fig:isic18_ex}
\end{figure}

\subsection{Datasets}
\label{sec:dataset}
We evaluated the performance of the proposed methods using two publicly available datasets, ISIC-18~\citep{Codella2019} and CIFAR-100~\citep{Krizhevsky2009}.
ISIC-18 is a typical medical image dataset with small number of classes and similar appearance between classes, making it suitable for evaluating the effectiveness of TTMA-DU and the class confusion characteristics of TTMA-CSU.
On the other hand, CIFAR-100 is a typical natural image dataset with a large number of diverse object classes, making it suitable for demonstrating the class similarity characteristics of TTMA-CSU.

\noindent
\textbf{ISIC-18~\citep{Codella2019}:} This dataset consists of 10,208 skin lesion images labeled with seven disease classes. 
The dataset is split into 10,015 images for training and 193 images for validation. 
As illustrated in Fig.~\ref{fig:isic18_ex}, although different classes exhibit common features, such as oval-shaped dark dots, there's a marked variance in the detailed appearance of skin lesions within individual classes.
These characteristics make ISIC-18 an ideal benchmark for evaluating the efficiency of TTMA-DU as aleatoric uncertainty and understanding class confusion characteristics of TTMA-CSU.

\noindent
\textbf{CIFAR-100~\citep{Krizhevsky2009}:} This dataset consists of 60,000 images labeled with 100 object classes.
The dataset is split into 50,000 images for training and 10,000 images for validation.
The dataset comprises a vast array of classes, each of them representing diverse types of natural scenes and objects. 
Its heterogeneity in terms of class variety makes it a suitable benchmark for investigating class similarity characteristics of TTMA-CSU.
In the experiment, 1,000 images were randomly selected from the 10,000 original validation images for validation, with 10 images for each class.

\subsection{Implementation details}
\label{sec:implementation}

All the experiments were developed on python and the PyTorch with eight NVIDIA RTX 2080 Ti GPU machines.

\noindent
\textbf{ISIC-18:} A VGG-19 model~\citep{Simonyan2014} was trained on the training set for 300 epochs using a mini-batch size of 128. 
The initial learning rate was set to 0.01 and was decreased by 10 after 150 and 225 epochs. 
Both affine and mixup data augmentation were applied during the training process. 
The affine data augmentation included a random horizontal flip, a random vertical flip, a random rotation between -45 and 45 degrees, a random translation with shift rates of (0.1,0.1), and a random scaling with a factor of 1 to 1.2.
In mixup augmentation, the mixup hyper-parameter $\alpha$ was set to 0.2. 
The drop-out probability for fully connected layers was set to 0.5. 

During the testing phase, augmentation methods suitable for each case, TTA, MCDO, and TTMA, were applied to the test data using the same parameters used during the training process. 
For TTA, affine augmentation was applied to the test data. 
For MCDO, drop-out was applied to the trained network on fully connected layers with a drop-out probability of 0.5. 
For TTMA, mixup augmentation with $\alpha=0.2$ was applied to the test data, where the number of selected training data for each class to compute the mixup was set to $K=30$. 
This resulted in $MK=210$ mixup augmented test data per one original test data. 

\noindent
\textbf{CIFAR-100:} A Wide Residual Network (WRN-28-10) model~\citep{Zagoruyko2016} was trained on the training set for 200 epochs using a mini-batch size of 256. 
The initial learning rate was set to 0.1 and was decreased by 5 after 60, 120, and 160 epochs. 
Both affine and mixup data augmentation were applied during the training process. 
The affine data augmentation included random cropping with a square size of 32, a random horizontal flip, a random rotation between -45 and 45 degrees, a random translation with shift rates of (0.1,0.1), and a random scaling with a factor of 1 to 1.2. 
In mixup augmentation, the mixup hyper-parameter $\alpha$ was set to 0.2. 
The drop-out probability for fully connected layers was set to 0.3. 

During the testing phase, augmentation methods suitable for each case, TTA, MCDO, and TTMA, were applied to the test data using the same parameters used during the training process. 
For TTA, affine augmentation was applied to the test data. 
For MCDO, drop-out was applied to the trained network on fully connected layers with a drop-out probability of 0.3. 
For TTMA, mixup augmentation with $\alpha=0.2$ was applied to the test data, where the number of selected training data for each class to compute the mixup was set to $K=10$. 
This resulted in $MK=1,000$ mixup augmented test data per original test data.

\begin{figure}[!t]
    \centering
        \subfloat[TTA]{\includegraphics[width=0.5\columnwidth]{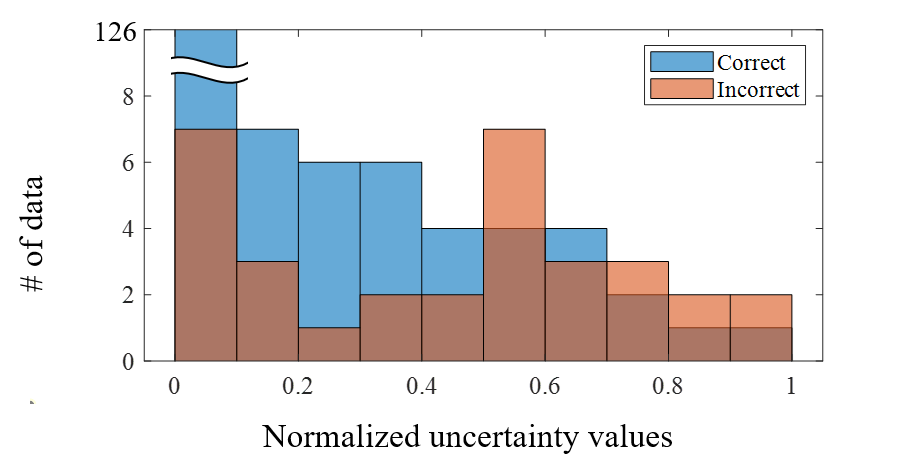}}
        \subfloat[MCDO]{\includegraphics[width=0.5\columnwidth]{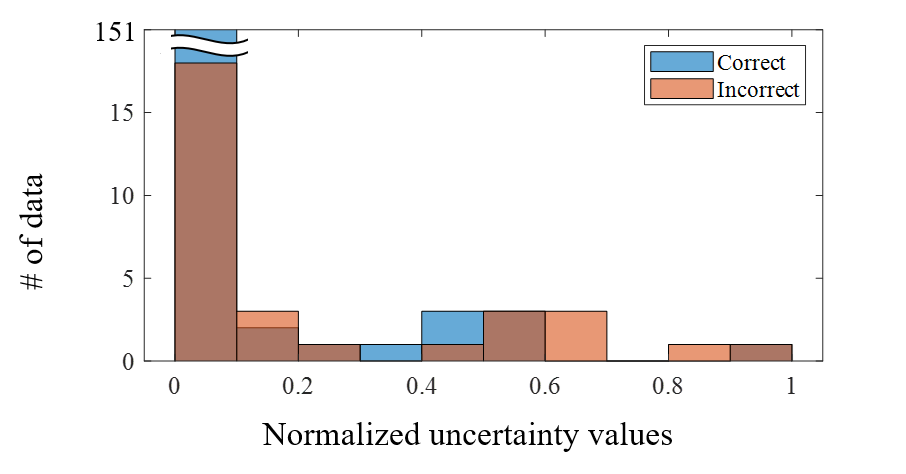}}
        \qquad
        \subfloat[TTMA-DU]{\includegraphics[width=0.5\columnwidth]{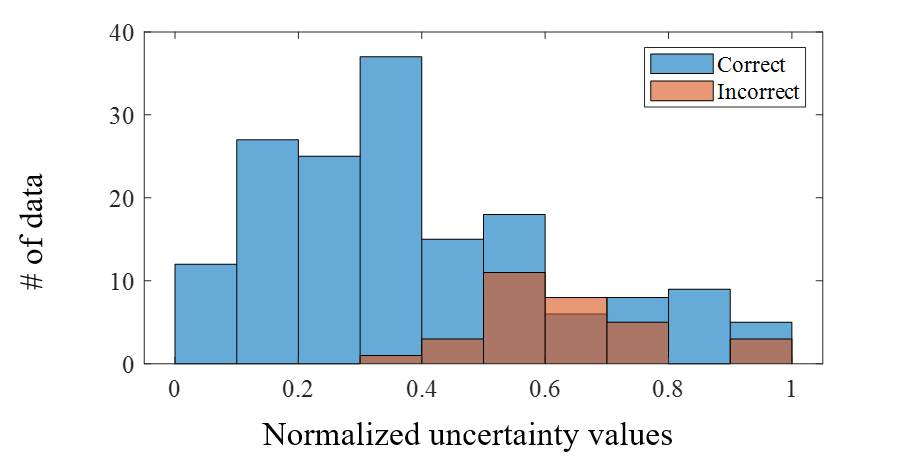}}
    \caption{Histograms of aleatoric uncertainty for correct and incorrect test data for ISIC-18 classification results with (a) TTA, (b) MCDO, and (c) TTMA-DU methods. In (a) TTA, the number of correct samples having uncertainty of [0,0.1) is 126. In (b) MCDO, the number of correct samples having uncertainty of [0,0.1) is 151.}
    \label{fig:hist_isic18}
\end{figure}

\begin{figure}[!t]
    \centering
        \subfloat[]{\includegraphics[width=0.5\columnwidth]{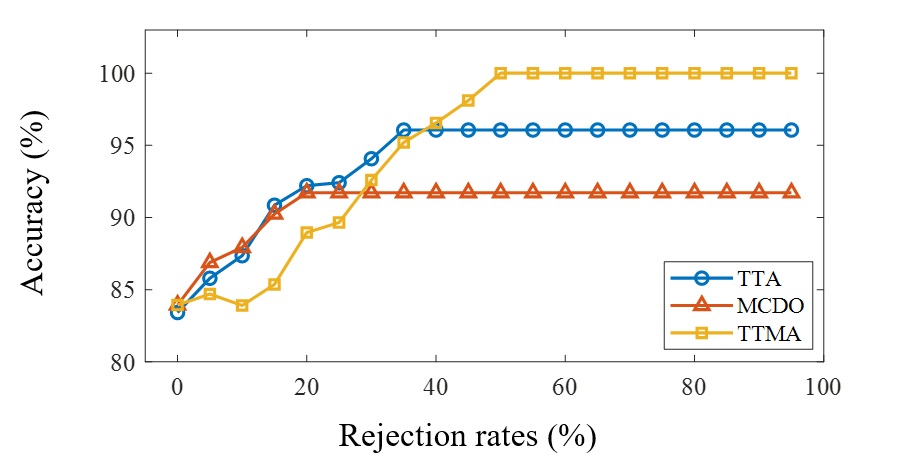}}
        \subfloat[]{\includegraphics[width=0.5\columnwidth]{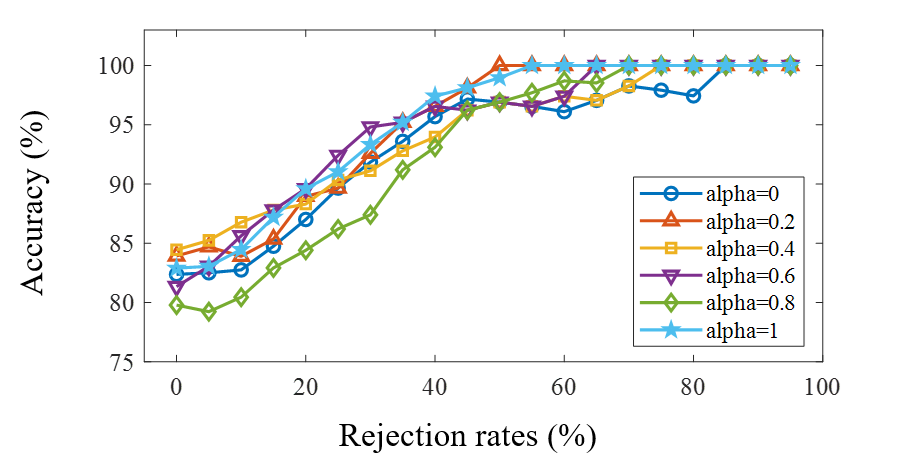}}
        \qquad
        \subfloat[]{\includegraphics[width=0.5\columnwidth]{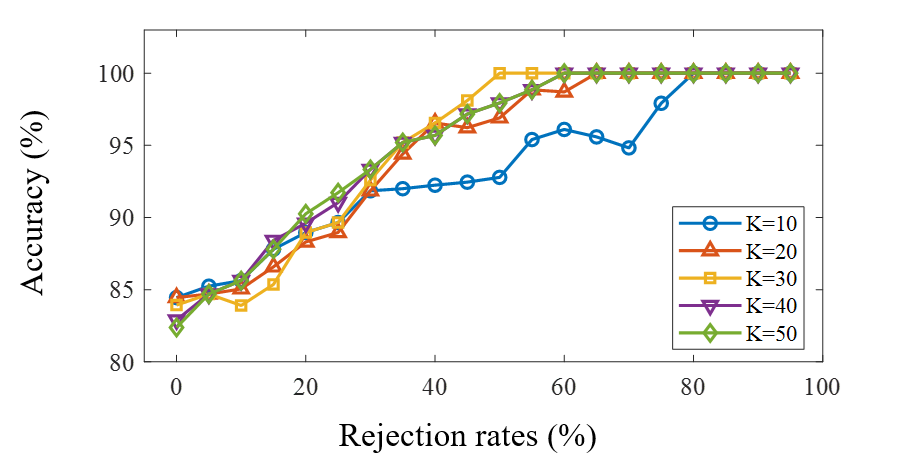}}
    \caption{Accuracy-rejection curves of ISIC-18 skin lesion classification results with (a) different uncertainty estimation methods, (b) different values of mixup hyper-parameter $\alpha$, and (c) different values of mixup sampling number $K$ for TTMA-DU.}
    \label{fig:arc_isic18}
\end{figure}

\begin{table}[!b]
\caption{Performance evaluation and comparisons of ISIC-18 skin lesion classification with different uncertainty estimation methods. Accuracy (\%) was evaluated on the rejected test data for various rejection rates $T\in\{0\%,25\%,50\%,75\%,95\%\}$.}
\centering
\resizebox{0.8\columnwidth}{!}{
\begin{tabular}{C{5cm} | C{1cm} C{1cm} C{1cm} C{1cm} C{1cm} }
\hline
 & \multicolumn{5}{c}{\textbf{Rejection rates}} \\
\textbf{Methods} & \textbf{0} & \textbf{25} & \textbf{50} & \textbf{75} & \textbf{95} \\
\hline \hline
Single & 83.4 &&&&  \\
\hline
TTA & 83.4 & \textbf{92.4} & 96.1 & 96.1 & 96.1  \\
MCDO & 83.9 & 91.7 & 91.7 & 91.7 & 91.7  \\
\hline
TTMA-DU ($\alpha=0.0$) & 82.4 & 89.7 & 96.9 & 97.9 & \textbf{100}  \\
TTMA-DU ($\alpha=0.2$) & 83.9 & 89.7 & \textbf{100} & \textbf{100} & \textbf{100}  \\
TTMA-DU ($\alpha=0.4$) & \textbf{84.5} & 90.3 & 96.9 & \textbf{100} & \textbf{100}  \\
TTMA-DU ($\alpha=0.6$) & 81.3 & \textbf{92.4} & 96.9 & \textbf{100} & \textbf{100}  \\
TTMA-DU ($\alpha=0.8$) & 79.8 & 86.2 & 96.9 & \textbf{100} & \textbf{100}  \\
TTMA-DU ($\alpha=1.0$) & 82.9 & 91.0 & 99.0 & \textbf{100} & \textbf{100}  \\
\hline
\end{tabular}
}
\label{table:result_isic18}
\end{table}

\subsection{Experiments on TTMA Data Uncertainty}
\label{sec:exp_du}

\subsubsection{Evaluation metrics and baseline methods}
To evaluate the efficiency of the proposed TTMA-DU as aleatoric measure, we examined uncertainty histograms for both correct and incorrect test data alongside accuracy-rejection curves.

\noindent
\textbf{Uncertainty histogram:} These histograms facilitate a comparison of the uncertainty distributions associated with correct and incorrect predictions. 
The aleatoric uncertainty measure is more effective when the uncertainty distributions for correct and incorrect predictions are more distinguishable.

\noindent
\textbf{Accuracy-rejection curves:} These curves plot the correlation between the rejection rate and accuracy when excluding test data that falls within the top $T$ uncertainty levels, where $T$ ranges from 0 to 95\%.
A more monotonically increasing and steeper curve indicates better discrimination of correct and incorrect predictions.

\noindent
\textbf{Baseline methods:} We compared TTMA-DU to two conventional aleatoric uncertainty methods: TTA~\citep{Wang2019} and MCDO~\citep{Gal2016}.
We also analyzed the sensitivity of TTMA-DU's hyperparameters $\alpha$ (mixup weight hyperparameter) and $K$ (number of selected training data per class for mixup).
We performed these comparisons using the VGG-19 model trained on ISIC-18 ($\alpha\in\{0,0.2,0.4,0.6,0.8,1\}$ and $K\in\{10,20,30,40,50\}$), and the WRN-28-10 model trained on CIFAR-100 ($\alpha\in\{0,0.2,0.4,0.6,0.8,1\}$ and $K\in\{5,10,15,20\}$).

\subsubsection{Results on TTMA-DU: ISIC-18}
Fig.~\ref{fig:hist_isic18} shows histograms of the aleatoric uncertainty for correct and incorrect test data for TTA, MCDO, and the proposed TTMA-DU. 
The ideal histogram would have two well-separated distributions, with correct test data concentrated in the low-uncertainty area and incorrect test data concentrated in the high-uncertainty area. 
However, the TTA and MCDO histograms have considerable overlap, making it difficult to distinguish between correct and incorrect samples. 
In contrast, the TTMA-DU histogram shows more distinguishable distributions. 
In addition, we can obtain 100\% accurate predictions by thresholding the test daa with a TTMA-DU value less than 0.3.

Fig.~\ref{fig:arc_isic18} shows the accuracy-rejection curves for different uncertainty estimation methods, different values of $\alpha$, and different values of $K$ using the proposed TTMA-DU. 
Table~\ref{table:result_isic18} summarizes the rejection accuracy for the proposed and the baseline methods.
In Fig.~\ref{fig:arc_isic18} (a), TTA reaches a maximum accuracy of 96.1\% at 35\% rejection, while MCDO reaches a maximum accuracy of 91.7\% with 20\% rejection. 
The proposed TTMA has a monotonically increasing curve, starting with an accuracy of 83.9\% at 0\% rejection and reaching 100\% accuracy at 50\% rejection.
In Fig.~\ref{fig:arc_isic18} (b), it can be seen that TTMA with a mixup weight parameter of $\alpha$=0.2 reaches 100\% accuracy most rapidly at 50\% rejection, compared to the other values of $\alpha$. 
Similarly, in Fig.~\ref{fig:arc_isic18} (c), it can be observed that TTMA with $K$=30 reaches 100\% accuracy most rapidly at 50\% rejection, compared to the other values of $K$.
Overall, the results confirm that the proposed TTMA-DU outperforms existing TTA and MCDO methods in distinguishing correct predictions from incorrect predictions.

\begin{figure}[!t]
    \centering
        \subfloat[TTA]{\includegraphics[width=0.5\columnwidth]{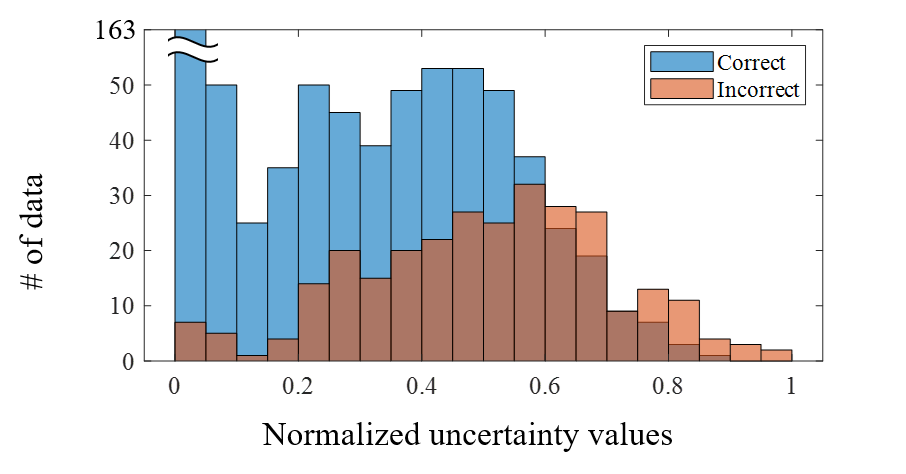}}
        \subfloat[MCDO]{\includegraphics[width=0.5\columnwidth]{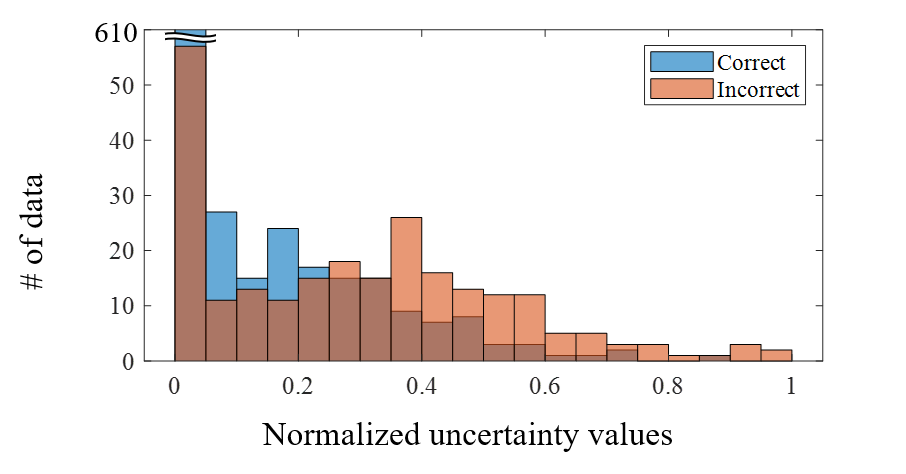}}
        \qquad
        \subfloat[TTMA-DU]{\includegraphics[width=0.5\columnwidth]{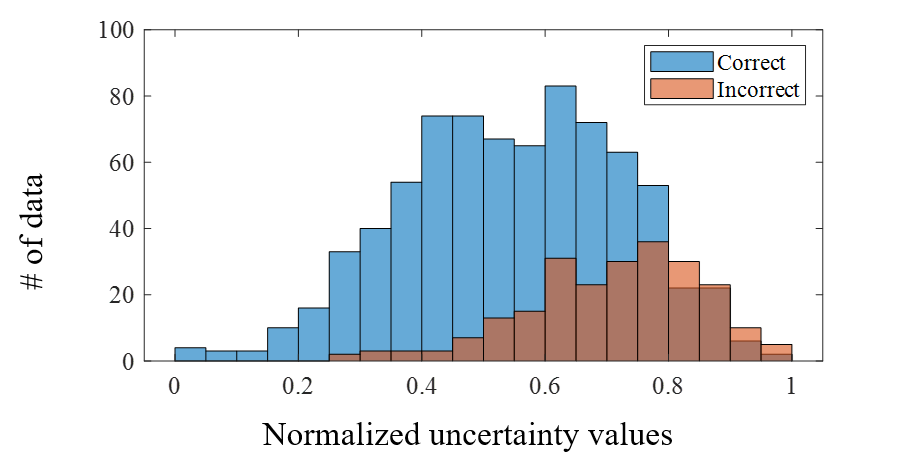}}
    \caption{Histograms of aleatoric uncertainty for correct and incorrect test data for CIFAR-100 classification results with (a) TTA, (b) MCDO, and (c) TTMA-DU methods. In (a) TTA, the number of correct samples having uncertainty of [0,0.05) is 163. In (b) MCDO, the number of correct samples having uncertainty of [0,0.05) is 610.}
    \label{fig:hist_cifar100}
\end{figure}

\begin{figure}[!t]
    \centering
        \subfloat[]{\includegraphics[width=0.5\columnwidth]{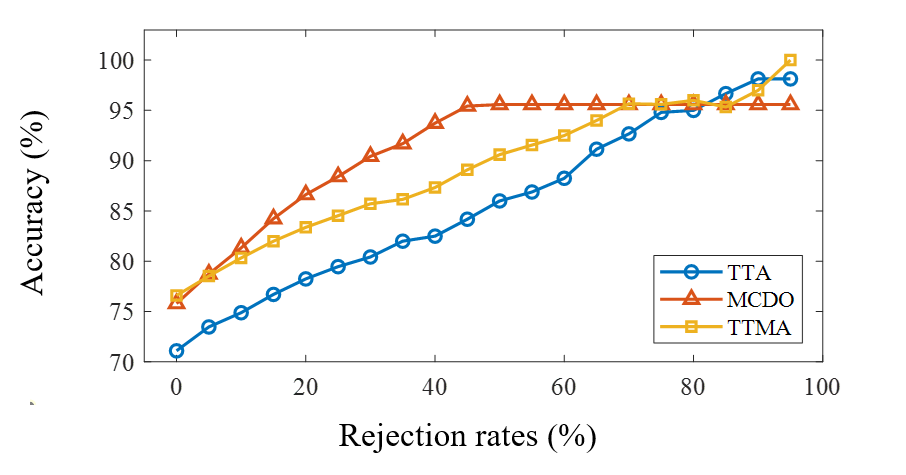}}
        \subfloat[]{\includegraphics[width=0.5\columnwidth]{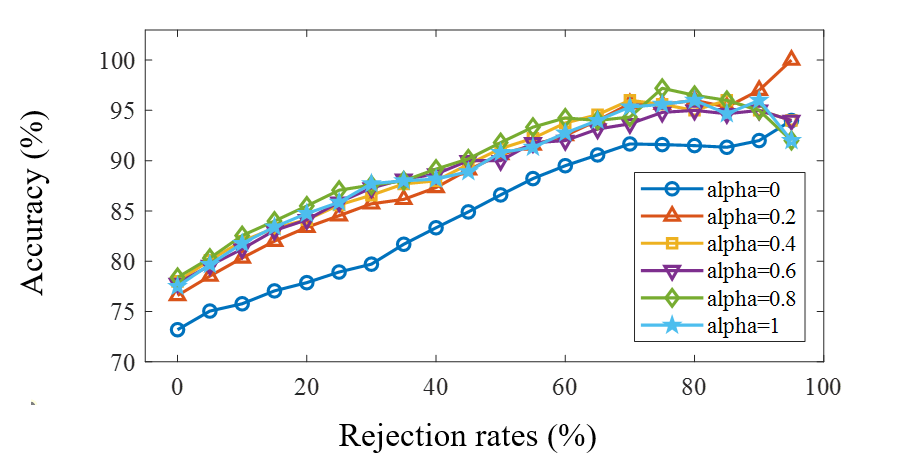}}
        \qquad
        \subfloat[]{\includegraphics[width=0.5\columnwidth]{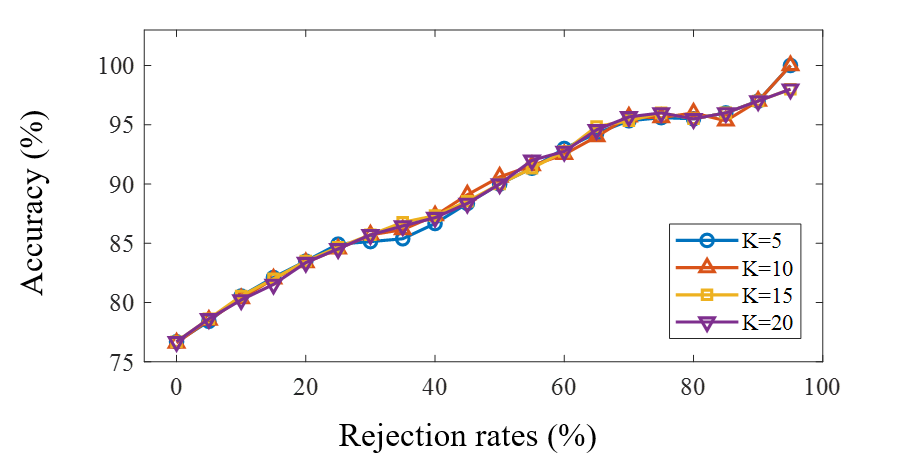}}
    \caption{Accuracy-rejection curves of CIFAR-100 classification results with (a) different uncertainty estimation methods, (b) different values of mixup hyper-parameter $\alpha$, and (c) different values of mixup sampling number $K$ for TTMA-DU.}
    \label{fig:arc_cifar100}
\end{figure}

\begin{table}[b!]
\caption{Performance evaluation and comparisons of CIFAR-100 classification with different uncertainty estimation methods. Accuracy (\%) was evaluated on the rejected test data for various rejection rates $T\in\{0\%,25\%,50\%,75\%,95\%\}$.}
\centering
\resizebox{0.8\columnwidth}{!}{
\begin{tabular}{C{5cm} | C{1cm} C{1cm} C{1cm} C{1cm} C{1cm} }
\hline
 & \multicolumn{5}{c}{\textbf{Rejection rates (\%)}} \\
\textbf{Methods} & \textbf{0} & \textbf{25} & \textbf{50} & \textbf{75} & \textbf{95} \\
\hline \hline
Single & 74.9 & & & & \\
\hline
TTA & 71.1 & 79.5 & 86.0 & 94.8 & 98.1  \\
MCDO & 75.8 & \textbf{88.4} & \textbf{95.6} & 95.6 & 95.6 \\
\hline
TTMA-DU ($\alpha=0.0$) & 73.2 & 78.9 & 86.6 & 91.6 & 94.0  \\
TTMA-DU ($\alpha=0.2$) & 76.6 & 84.5 & 90.6 & 95.6 & \textbf{100}  \\
TTMA-DU ($\alpha=0.4$) & 78.0 & 85.6 & 91.2 & 95.6 & 94.0 \\
TTMA-DU ($\alpha=0.6$) & 77.8 & 85.9 & 90.0 & 94.8 & 94.0 \\
TTMA-DU ($\alpha=0.8$) & \textbf{78.4} & 87.1 & 91.8 & \textbf{97.2} & 92.0  \\
TTMA-DU ($\alpha=1.0$) & 77.5 & 85.9 & 90.8 & 95.6 & 92.0  \\
\hline
\end{tabular}
}
\label{table:result_cifar100}
\end{table}

\subsubsection{Results on TTMA-DU: CIFAR-100}
Fig.~\ref{fig:hist_cifar100} shows histograms of the aleatoric uncertainty for correct and incorrect test data for TTA, MCDO, and the proposed TTMA-DU.
In both TTA and MCDO, correct and incorrect test data range the spectrum from the lowest to the highest uncertainty values, making it difficult to distinguish between correct and incorrect predictions with these uncertainty measures.
In contrast, TTMA-DU shows almost no incorrect test data in the lowest uncertainty area and a shift of the overall distribution to the higher uncertainty area, making it more distinguishable compared to the other methods.
Moreover, we can extract 100\% accurate predictions by selecting the test data with a TTMA-DU value less than 0.2.

Fig.~\ref{fig:arc_cifar100} shows the accuracy-rejection curves for (a) different uncertainty estimation methods, (b) different values of $\alpha$, and (c) different values of $K$ with the proposed TTMA-DU method. 
Table~\ref{table:result_cifar100} summarizes the rejection accuracy for the proposed and the baseline methods.
In Fig.~\ref{fig:arc_cifar100} (a), TTA shows a curve that increases steeply from a low 0\% rejection accuracy, while MCDO shows a curve that saturates at 50\% rejection accuracy. 
In contrast, the proposed TTMA-DU method starts with a similar 0\% rejection accuracy to MCDO, but shows a curve that increases monotonically as the rejection rate increases, achieving the highest 95\% rejection accuracy.
In Fig.~\ref{fig:arc_cifar100} (b), it can be observed that TTMA-DU with a mixup weight parameter $\alpha$=0.2 achieves the highest 95\% rejection accuracy compared to other choices of $\alpha$.
Fig.~\ref{fig:arc_cifar100} (c) indicates that the accuracy-rejection characteristics of the proposed TTMA-DU are relatively less affected by the number of mixup samples $K$.


\subsection{Experiments on TTMA Class-Specific Uncertainty}
\label{sec:exp_csu}

\begin{figure}[!t]
    \centering
    \includegraphics[width=0.5\columnwidth]{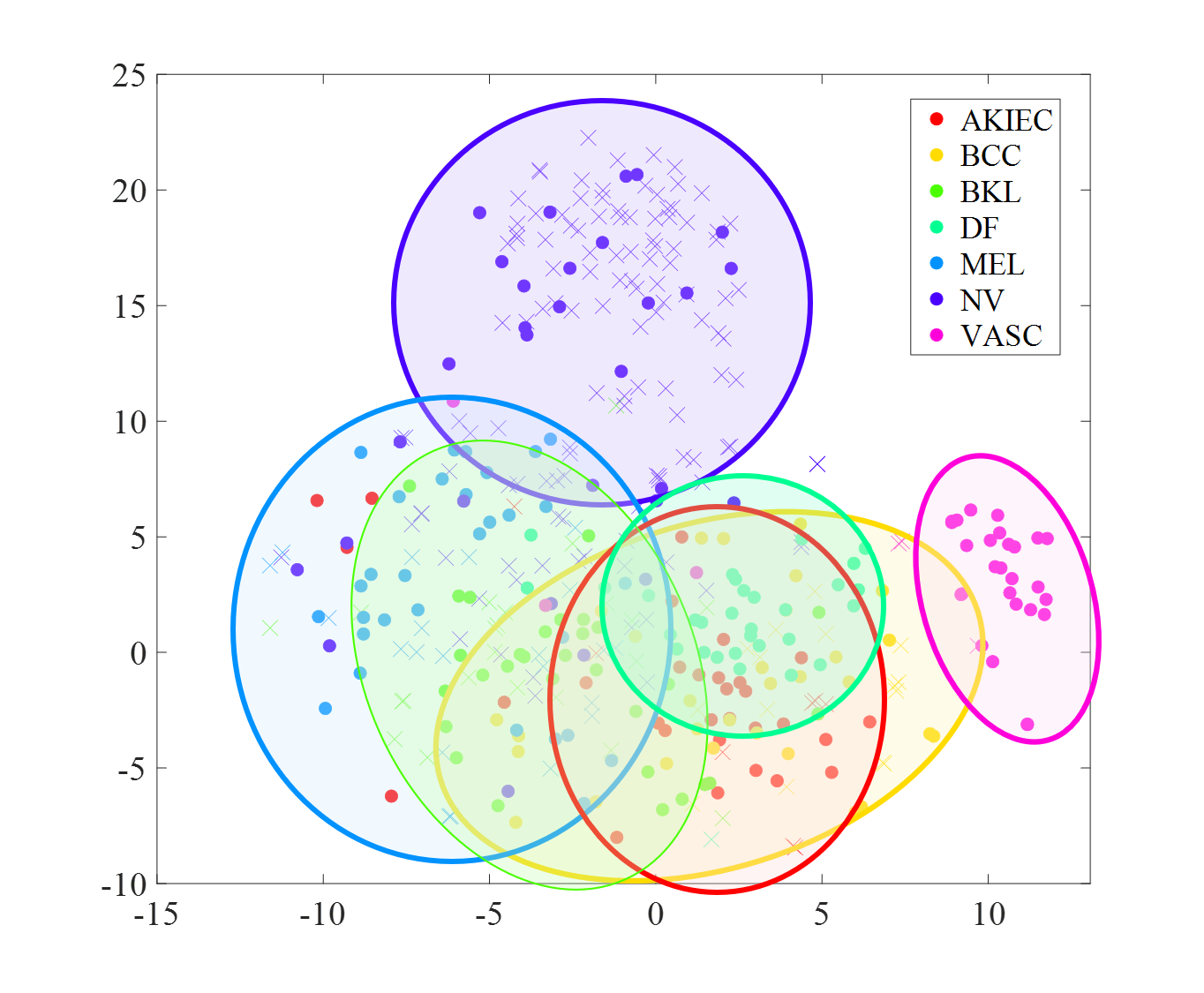}
    \caption{tSNE feature distributions of training (marked as o) and validation (marked as x) data in the ISIC-18 skin lesion classification results.}
    \label{fig:tsne_isic18}
\end{figure}

\subsubsection{Evaluation metrics and baseline methods}

To determine whether the proposed TTMA-CSU method provides information about class confusion and similarity, we compared TTMA-CSU for each class with the Average Feature Distance (AFD) in the feature space. 
We used boxplots to visualize the TTMA-CSU and AFD distributions, with feature distributions depicted through t-SNE (t-distributed stochastic neighbor embedding)~\citep{vanderMaaten2008} to verify if the proposed TTMA-CSU can help distinguishing class confusion from class similarity.

\begin{figure}[!t]
    \centering
        \subfloat[TTMA-CSU (AKIEC)]{\includegraphics[width=0.45\columnwidth]{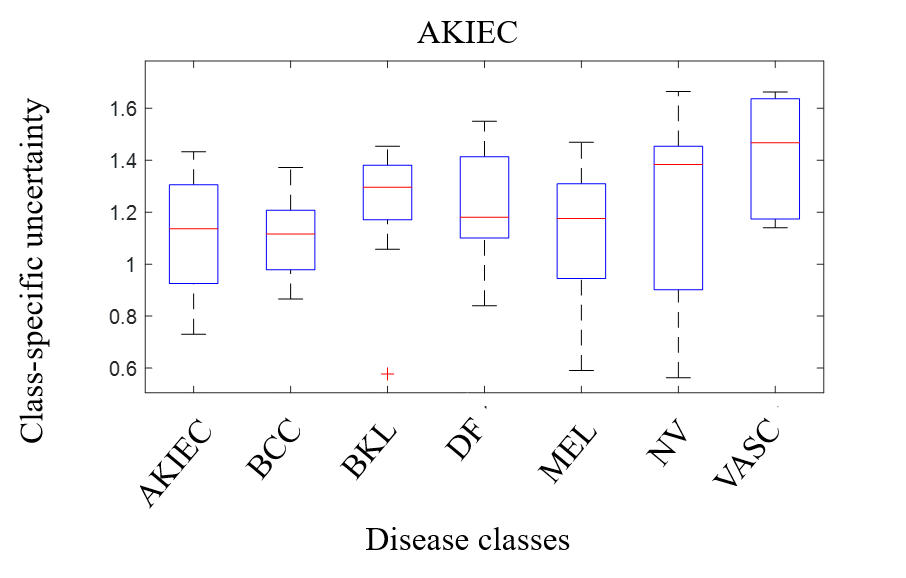}}
        \subfloat[AFD (AKIEC)]{\includegraphics[width=0.45\columnwidth]{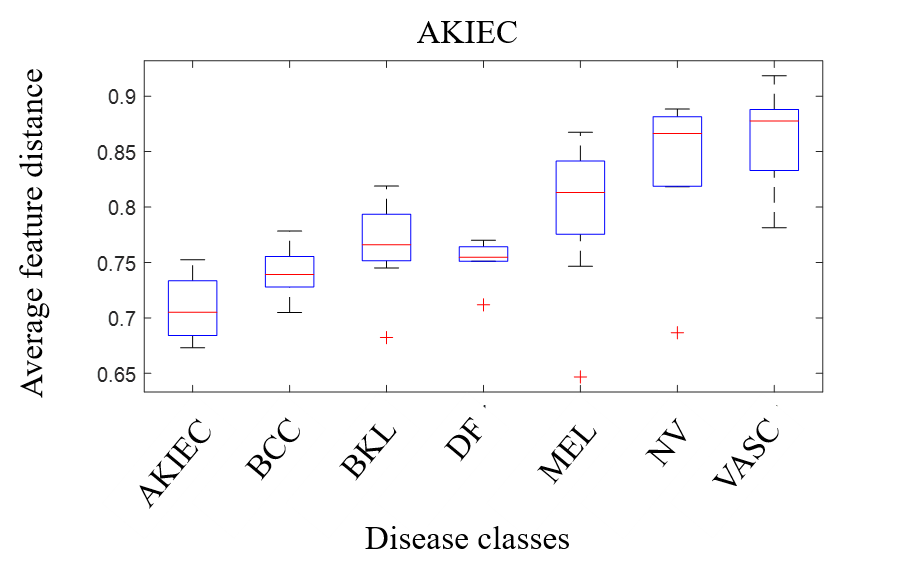}}
    \caption{Boxplots of (a) TTMA class-specific uncertainty (TTMA-CSU) and (b) average feature distance (AFD) for actinic keratosis (AKIEC) class in the ISIC-18 classification results. }
    \label{fig:csu_box_akiec}
\end{figure}

\begin{figure}[!t]
    \centering
        \subfloat[TTMA-CSU (MEL)]{\includegraphics[width=0.45\columnwidth]{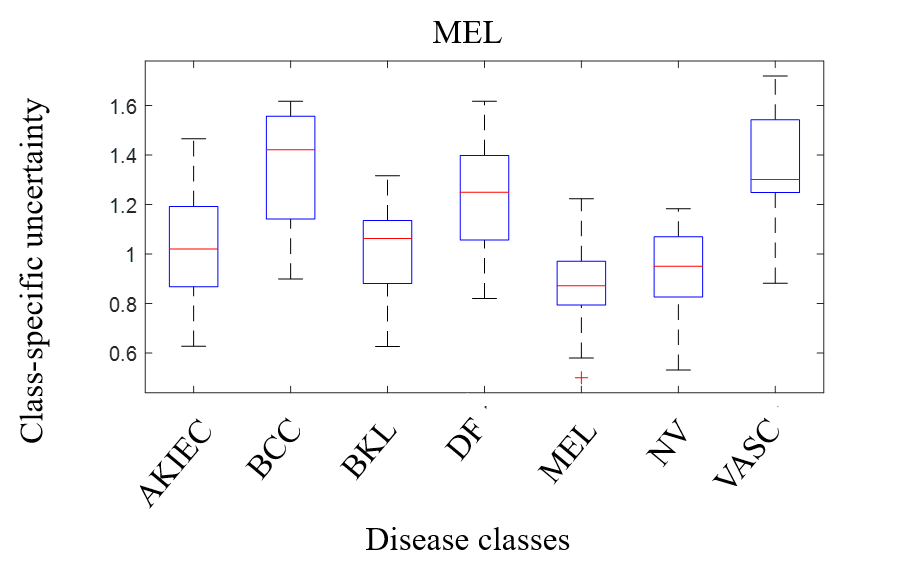}}
        \subfloat[AFD (MEL)]{\includegraphics[width=0.45\columnwidth]{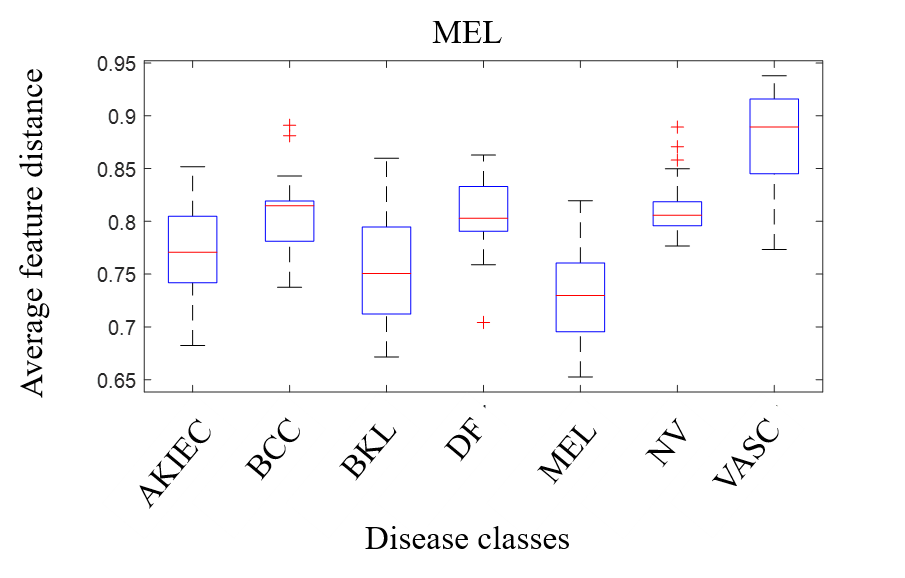}}
    \caption{Boxplots of (a) TTMA class-specific uncertainty (TTMA-CSU) and (b) average feature distance (AFD) for melanoma (MEL) class in the ISIC-18 classification results. }
    \label{fig:csu_box_mel}
\end{figure}

\subsubsection{Results on TTMA-CSU: ISIC-18}

Fig.~\ref{fig:tsne_isic18} illustrates the t-SNE feature distributions of the sampled training and validation data of the ISIC-18 dataset. 
In latent space, the AKIEC cluster (red) exhibits substantial overlap with the clusters of BCC (yellow), BKL (light green), and DF (cyan).
It is also distributed near the clusters of MEL (blue) and NV (indigo), while maintaining a distinct separation from the cluster of VASC (purple).
On the other hand, the MEL cluster (blue) overlaps significantly with the BCC (yellow), and BKL (light green), is disrtributed close to the NV (indigo), and remains distant from the VASC (purple).
Our analysis aims to assess whether the relationship between TTMA-CSU and AFD for these classes is consistent with these distribution characteristics observed in the latent space.

Figs.~\ref{fig:csu_box_akiec} and \ref{fig:csu_box_mel} present box plots of TTMA-CSU and AFD for various classes specifically for test data from the AKIEC and MEL classes, respectively.
In Fig.~\ref{fig:csu_box_akiec}, it can be observed that BKL, BCC, and DF, which overlap with AKIEC, exhibit relatively low AFD while maintaining relatively high TTMA-CSU.
This aligns with our hypothesis that in scenarios of class confusion, AFD tends to be lower, but TTMA-CSU increases due to the instability of prediction of mixup data.
In addition, MEL and NV, which are close to AKIEC, display relatively high AFD while maintaining relatively low TTMA-CSU.
This suggests that the closeness of two classes leads to more stable mixup predictions, thus lowering TTMA-CSU, as mixup helps in delineation of clearer boundaries between these classes.
Conversely, VASC, which is distant from AKIEC, exhibits both high TTMA-CSU and AFD.
This indicates that when two classes are heterogeneous, mixup acts as a heavy perturbation, destabilizing the predictions and consequently increasing TTMA-CSU.
Fig.~\ref{fig:csu_box_mel} illustrates that BCC and BKL, which overlap with MEL, exhibit relatively low AFD while maintaining relatively high TTMA-CSU.
NV, which is close to MEL, displays lower TTMA-CSU while maintaining relatively high AFD.
On the other hand, VASC, which is distant from MEL, exhibits both high TTMA-CSU and AFD.
These observations suggest that TTMA-CSU and AFD provide insights not only into the distance between classes in latent space, but also into the classifier's effectiveness in distinguishing and confusing between various classes.

\begin{figure}[!t]
    \centering
    \includegraphics[width=0.5\columnwidth]{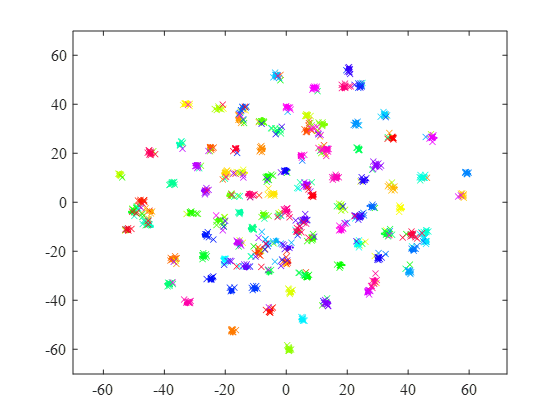}
    \caption{tSNE feature distributions of training (marked as o) and validation (marked as x) data in the CIFAR-100 classification results.}
    \label{fig:tsne_cifar100}
\end{figure}

\begin{figure*}[!t]
    \centering
        \subfloat[]{\includegraphics[width=0.06\textwidth]{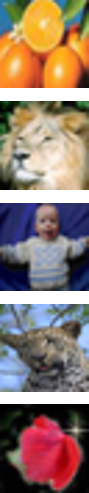}}
        \hspace{0.3cm}
        \subfloat[]{\includegraphics[width=0.06\textwidth]{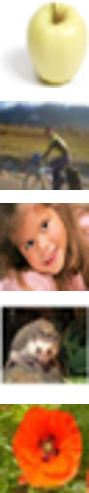}}
        \subfloat[]{\includegraphics[width=0.06\textwidth]{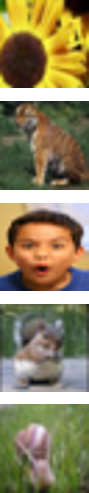}}
        \subfloat[]{\includegraphics[width=0.06\textwidth]{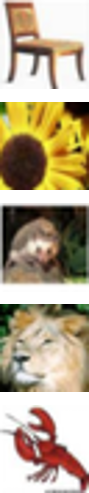}}
        \subfloat[]{\includegraphics[width=0.06\textwidth]{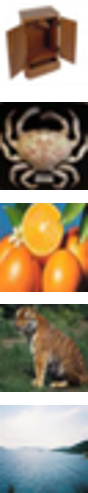}}
        \subfloat[]{\includegraphics[width=0.06\textwidth]{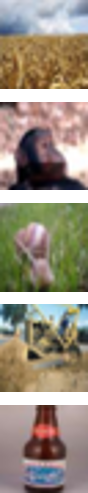}}
        \hspace{0.3cm}
        \subfloat[]{\includegraphics[width=0.06\textwidth]{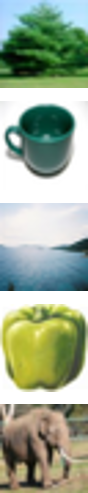}}
        \subfloat[]{\includegraphics[width=0.06\textwidth]{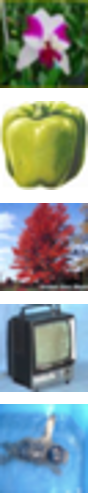}}
        \subfloat[]{\includegraphics[width=0.06\textwidth]{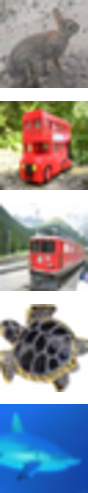}}
        \subfloat[]{\includegraphics[width=0.06\textwidth]{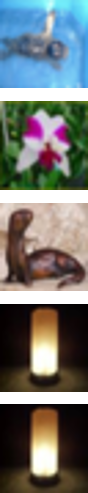}}
        \subfloat[]{\includegraphics[width=0.06\textwidth]{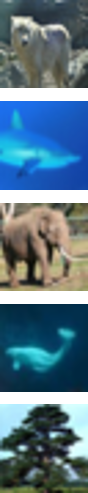}}
    \caption{Examples of CIFAR-100 test classes (a) and the classes with the lowest five TTMA-CSU (b-f) and the classes with the highest five TTMA-CSU (g-k). Five example test classes in (a) are orange, lion, baby, leopard, and rose (top to bottom).}
    \label{fig:bccsu_ex_cifar100}
\end{figure*}

\begin{figure*}[!t]
    \centering
        \subfloat[TTMA-CSU (Orange)]{\includegraphics[width=\textwidth]{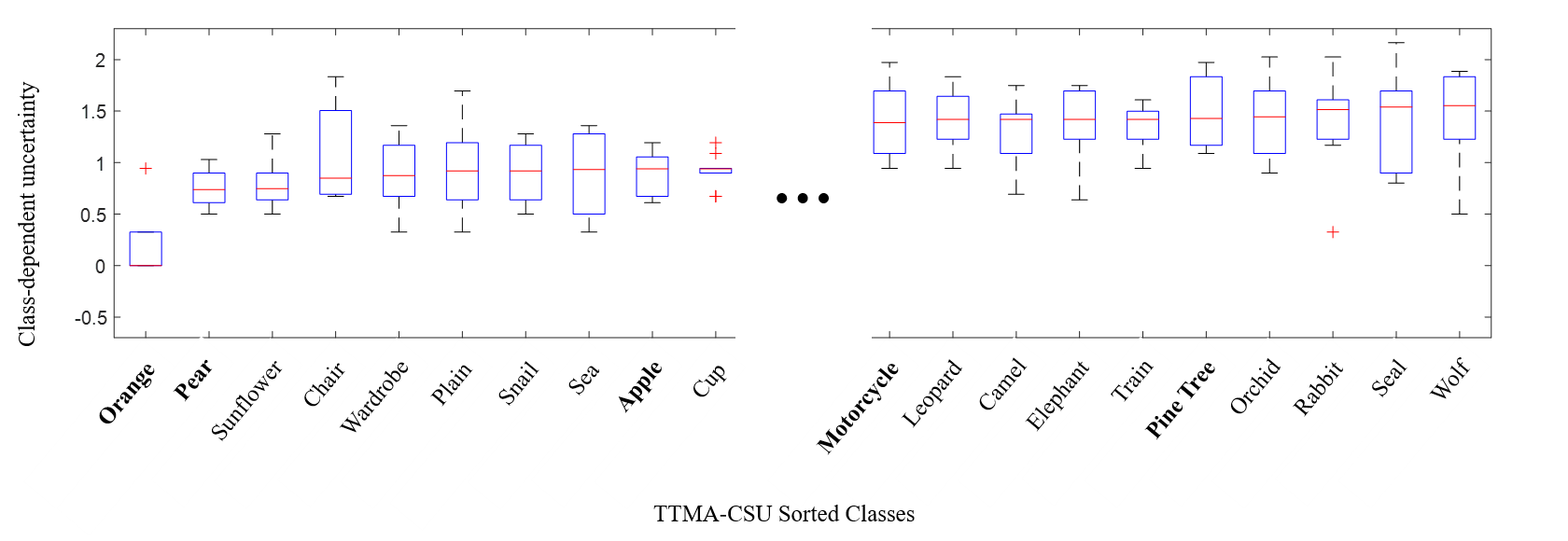}}
        \qquad
        \subfloat[AFD (Orange)]{\includegraphics[width=\textwidth]{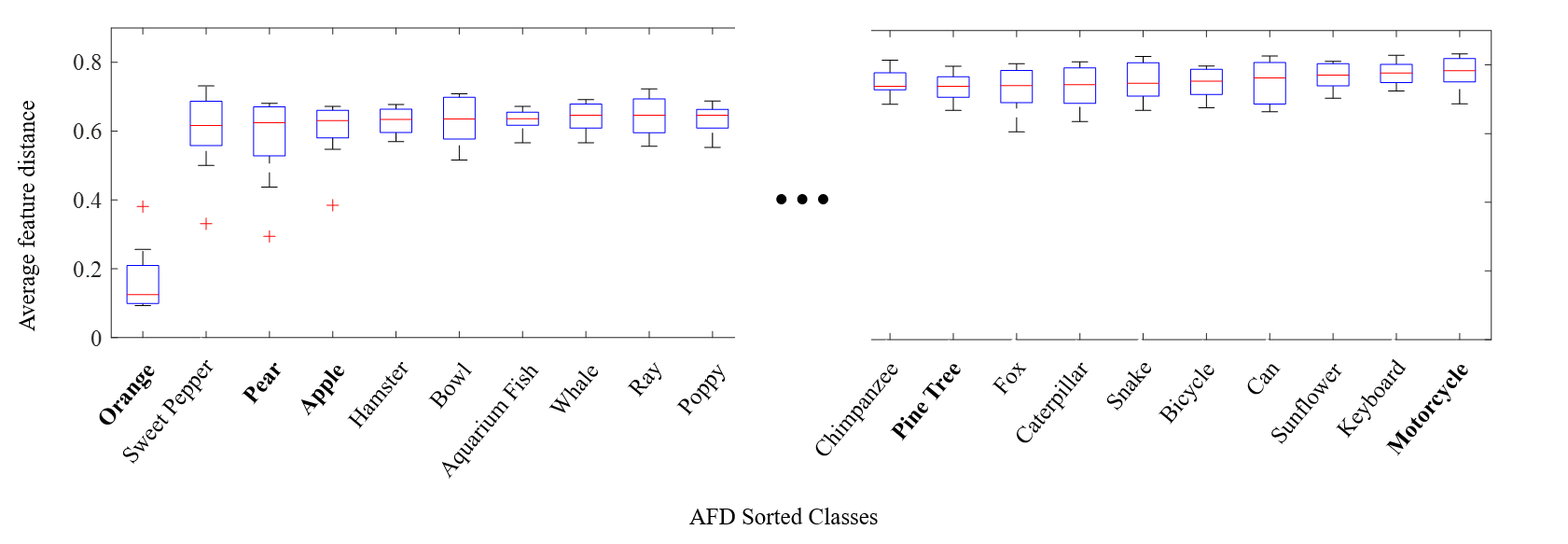}}
    \caption{Boxplots of (a) TTMA class-specific uncertainty (TTMA-CSU) and (b) average feature distance (AFD) for ''orange`` class in the CIFAR-100 classification results. Classes are sorted according to the median values. The ten classes with the lowest medians and the ten classes with the highest medians are shown on the left and right sides, respectively. The classes corresponding to both of the lowest/highest ten in-class TTMA-CSU and in-class AFD are shown in bold.}
    \label{fig:bccsu_box_cifar100}
\end{figure*}

\subsubsection{Results on TTMA-CSU: CIFAR-100}

Fig.~\ref{fig:tsne_cifar100} illustrates tSNE feature distributions of the sampled training and validation data in CIFAR-100. 
Unlike ISIC-18, the feature distribution of CIFAR-100 has almost no overlap between the classes, and the data of each class are densely distributed.
This implies that it can be expected that TTMA-CSU will provide only information about class similarity, rather than class confusion.

Fig.\ref{fig:bccsu_ex_cifar100} presents the images of five low TTMA-CSU classes and five high TTMA-CSU classes for five CIFAR-100 test data. 
It can be observed that the classes with low TTMA-CSU (1) belong to the same super-class, e.g., lion-tiger and baby-boy, or (2) belong to different classes but have similar appearances in terms of color and texture, e.g., orange-sunflower, lion-crab, and rose-lobster. 
In contrast, the high TTMA-CSU classes include classes with different super-classes and appearances, e.g., lion-cup, baby-train, and rose-shark. 
Fig.\ref{fig:bccsu_box_cifar100} shows boxplots of (a) TTMA-CSU and (b) AFD for the "orange" class. 
It can be observed that the low TTMA-CSU classes include (1) classes of the same super-class, e.g., apple and pear, and (2) classes with similar color appearances, e.g., sunflower and chair. 
On the other hand, the low AFD classes include (1) classes of the same super-class, e.g., apple and pear, and (2) classes of a similar round shape, e.g., sweet pepper, hamster, and bowl.
This confirms that the proposed TTMA-CSU provides information on class similarity in a similar manner to AFD for a dataset without class confusion.

\section{Discussion}
\label{sec:discussion}

Evaluating the trustworthiness of network predictions is a crucial aspect of uncertainty estimation. 
It involves distinguishing between correct and incorrect predictions based on the estimated uncertainty without relying on ground truth labels. 
An ideal uncertainty estimation method should assign low uncertainty values to all correct predictions and high uncertainty values to all incorrect predictions. 
However, convolutional neural networks (CNNs) are known for their inherent overconfidence, resulting in low uncertainty values for most decisions. 
To address this issue, it is essential to lower the network's confidence and calibrate it by applying perturbations to the input data or network or by employing ensemble methods for uncertainty estimation. 
Both test-time augmentation (TTA) and Monte Carlo dropout (MCDO) utilize perturbations and ensemble techniques, such as affine augmentation of data and dropout on the network, respectively. 
However, as demonstrated in the experimental results for both datasets, these methods do not fully alleviate the low uncertainty issue. 
This motivated us to explore the possibility of employing a stronger perturbation to achieve a more accurate uncertainty measure. 
As anticipated, the proposed TTMA-DU exhibits enhanced discrimination between correct and incorrect predictions compared to conventional TTA or MCDO.

\begin{table}[b!]
\caption{Expected calibration error (ECE) evaluation and comparison of ISIC-18 and CIFAR-100 classification results with different uncertainty estimation methods.}
\centering
\resizebox{0.5\columnwidth}{!}{
\begin{tabular}{C{3cm} | C{2.5cm} C{2.5cm} }
\hline
 & \multicolumn{2}{c}{\textbf{Datasets}} \\
\textbf{Methods} & \textbf{ISIC-18}  & \textbf{CIFAR-100} \\
\hline \hline
Single  & \textbf{0.0729}  & \textbf{0.0992} \\
\hline
TTA & 0.0878  & 0.1475 \\
MCDO & 0.1316  & 0.1638 \\
\hline
TTMA ($\alpha=0.0$) & 0.1686 & 0.3733 \\
TTMA ($\alpha=0.2$) & 0.2196 & 0.3641 \\
TTMA ($\alpha=0.4$) & 0.1663 & 0.3687 \\
TTMA ($\alpha=0.6$) & 0.1529 & 0.3586 \\
TTMA ($\alpha=0.8$) & 0.0958 & 0.3538 \\
TTMA ($\alpha=1.0$) & 0.1750 & 0.3378 \\
\hline
\end{tabular}
}
\label{table:ece}
\end{table}

To further investigate the reasons behind TTMA-DU's superior performance in distinguishing between correct and incorrect predictions compared to conventional TTA or MCDO, we analyze them from the perspective of network ensemble calibration. 
Table~\ref{table:ece} presents the expected calibration error (ECE) for the proposed and comparative methods on the ISIC-18 and CIFAR-100 datasets. 
ECE measures the difference between the predicted confidence and the actual accuracy~\citep{Naeini2015}, with a lower ECE indicating a better-calibrated network. 
As evident from Table~\ref{table:ece}, the proposed TTMA method exhibits higher ECEs than TTA and MCDO. 
This aligns with recent findings on the relationship between mixup augmentation, ensemble learning, and network calibration~\citep{Wen2021}. 
Wen~\textit{et al.} reported that mixup and ensemble have the effect of lowering the confidence of overconfident deep neural networks. 
When both mixup and ensemble are applied, the effects accumulate, leading to an under-confident network. 
This under-confidence characteristic of TTMA is also observable from the uncertainty distributions in Fig.\ref{fig:hist_isic18} and Fig.\ref{fig:hist_cifar100}. 
The proposed TTMA exhibits a smoother confidence distribution compared to those of TTA and MCDO, and the entropy-based uncertainty is also distributed at higher values than those of TTA and MCDO. 
This under-confidence characteristic of the proposed method enhances the distinction between correct and incorrect predictions by further increasing the uncertainty values of incorrect predictions, making it more challenging for the model to be overly confident in its incorrect predictions. 


In addition to the aleatoric uncertainty measure, TTMA-DU, we propose TTMA-CSU, a novel type of uncertainty measure that jointly considers data and class information. 
Unlike TTMA-DU, which mixes training data from all classes for uncertainty estimation, TTMA-CSU utilizes data from a specific class, revealing insights into prediction instability between the target data and that class.
We hypothesized that TTMA-CSU would offer the insights on class confusion and similarity, crucial aspects for analyzing trained networks. 
Traditional measures like AFD is not sufficient to identify these relationships, yielding low values for both class confusion and similarity. 
TTMA-CSU shows distinct behaviors for class confusion and similarity, enabling effective differentiation between them along with AFD.
Experiments comparing TTMA-CSU and AFD in t-SNE feature space confirmed our hypothesis. 
In confusion scenarios, where classes overlap in the latent space, TTMA-CSU exhibited high values alongside low AFD. 
Conversely, for class similarities with distinct but close representations, both TTMA-CSU and AFD remained low. 
It can be confirmed that the TTMA-CSU and AFD can quantify the class confusion and class similarity characteristics of the trained network, which can offer valuable insights for assessing the trained network's performance and identifying potential areas for improvement.


One limitation of the proposed method lies in its computational complexity.
While sampling all classes for mixup in TTMA works well for smaller datasets, it becomes computationally expensive for large-scale datasets e.g. ImageNet with 1,000 classes.
For instance, sampling five mixup samples from each class results in the necessity of conducting 5,000 mixups for a single test data.
To mitigate this complexity, future work can include strategies for selective sampling of classes for mixup in TTMA.
Randomly selecting mixup classes from all classes can introduce unnecessary bias and degrade the performance of the uncertainty measure.
One alternative strategy can involve dividing all classes into \textit{close} and \textit{distant} class groups from the test data using TTMA-CSU, and selecting mixup classes from each of these class groups.
This selective strategy can reduce computational complexity while preserving the reliability of the uncertainty measure through minimizing unnecessary bias.


\section{Conclusion}
\label{sec:conclusion}

In this paper, we introduced a TTMA-based uncertainty estimation method for deep learning classification. 
First, we proposed TTMA-DU which yields more intense perturbations compared to TTA and MCDO, leading to better evaluation of the trained network's trustworthiness.
Second, we introduced TTMA-CSU which is a novel type of uncertainty measure providing valuable insights into class confusion and class similarity within the trained network.
Experiments on two publicly available image classification datasets demonstrated that (1) the proposed TTMA-DU consistently outperforms existing uncertainty estimation methods in differentiating correct and incorrect predictions, and (2) the proposed TTMA-CSU shows the ability to distinguish class confusion from class similarity along with AFD.
Future work could focus on improving the computational efficiency of TTMA and investigating its applicability to other tasks beyond image classification.

\section*{Acknowledgments}
This work was supported by the National Research Foundation of Korea (NRF) grants funded by the Korea government (MSIT) (No. 2020R1A2C1102140 and RS-2023-00207947), and the Basic Science Research Program through the National Research Foundation of Korea (NRF) funded by the Ministry of Education (2022R1I1A1A01071970).

\bibliography{MAIN}
\bibliographystyle{iclr2021_conference}


\end{document}